\documentclass[conference]{IEEEtran}

\ifCLASSINFOpdf
 
\else
 
\fi

\usepackage{cite}
\usepackage{hyperref}
\usepackage[english]{babel}
\usepackage{graphicx,caption}
\usepackage{subfig}
\graphicspath{{Figures/}}
\DeclareGraphicsExtensions{.pdf,.jpeg,.png}
\usepackage{amsmath,amsfonts,amssymb,booktabs,threeparttable}
\usepackage{gensymb}
\usepackage{xcolor,colortbl}
\usepackage[mathlines,switch]{lineno}
\usepackage{multirow}

\definecolor{cerulean}{rgb}{0.0, 0.48, 0.65}
\definecolor{grn}{rgb}{0, .8, .6}
\definecolor{brickred}{rgb}{0.8, 0.25, 0.33}

\begin{document}

\title{Hourly-Similarity Based Solar Forecasting Using Multi-Model Machine Learning Blending}

\author{\IEEEauthorblockN{Cong~Feng, Jie~Zhang}
\IEEEauthorblockA{The University of Texas at  Dallas\\Richardson, TX, 75080 USA}}

\maketitle

\begin{abstract}
With the increasing penetration of solar power into power systems, forecasting becomes critical in power system operations. In this paper, an hourly-similarity (HS) based method is developed for 1-hour-ahead (1HA) global horizontal irradiance (GHI) forecasting. This developed method utilizes diurnal patterns, statistical distinctions between different hours, and hourly similarities in solar data to improve the forecasting accuracy. The HS-based method is built by training multiple two-layer multi-model forecasting framework (MMFF) models independently with the same-hour subsets. The final optimal model is a combination of MMFF models with the best-performed blending algorithm at every hour. At the forecasting stage, the most suitable model is selected to perform the forecasting subtask of a certain hour. The HS-based method is validated by 1-year data with six solar features collected by the National Renewable Energy Laboratory (NREL). Results show that the HS-based method outperforms the non-HS (all-in-one) method significantly with the same MMFF architecture, wherein the optimal HS-based method outperforms the best all-in-one method by 10.94\% and 7.74\% based on the normalized mean absolute error and normalized root mean square error, respectively.
\end{abstract}


\IEEEpeerreviewmaketitle

\section{Introduction}
With the intensify of energy consumption and environment deterioration, facilitating and integrating renewable energy become increasingly important. Solar energy has increased rapidly and the cost also continuously decreases in recent years~\cite{lewis2016research}. However, it is still challenging to reliably and economically operate power systems under large penetrations of solar energy due to the uncertainty and variability in solar power. Improving solar forecasting accuracy could partially address the challenges of foreseeable large solar penetrations faced by independent system operators and independent power producers.

For short-term solar forecasting, statistical methods are popularly used due to their learning power and high computational efficiency. Based on the algorithm complexity, statistical methods can be categorized into traditional time series methods (e.g., autoregressive integrated moving average (ARIMA) method~\cite{david2016probabilistic}), machine learning methods (e.g., support vector machine (SVM)~\cite{deo2016wavelet}), and deep learning models (e.g., long short term memory neural network (LSTM-NN)~\cite{gensler2016deep}). A more comprehensive review of statistical methods for solar forecasting can be found in recent review papers~\cite{voyant2017machine, antonanzas2016review, raza2016recent}.

To improve forecasting accuracy, different techniques have been proposed in the literature, such as dividing forecasting into different subtasks. For example, Feng et al.~\cite{feng1short} grouped forecasts by weather conditions and improved the forecasting performance by 20\%. Temperature was used in~\cite{ding2011ann} to determine a suitable forecasting model for the similar day to enhance the accuracy. Forecasting could be also differentiated by forecast error metrics, thereby generating superior forecasts~\cite{wu2013prediction}.

In an attempt to more precisely forecast short-term global horizontal irradiance (GHI), an hourly-similarity (HS) based method is developed in this study, using a two-layer hybrid machine learning based multi-model forecasting framework (MMFF). Contributions of this paper are twofold: (i) to develop an advanced forecasting method based on an HS classification criterion; (ii) to take advantage of MMFF's powerful learning ability. 

The remainder of the paper is organized as follows: Section II explores the hourly similarities in solar feature time series. The developed HS-based MMFF framework is described in Section III. An experiment study is used to validate the proposed method in Section IV. Section V concludes the paper.

\section{Hourly Similarities in Solar Data}
\subsection{Solar Features}
The forecasting performance of data-driven statistical models is highly influenced by their inputs. To obtain well-performing models, six solar features are extracted from two types of data source: i) GHI features: historical GHI ({$\boldsymbol{GHI}$}), clear sky GHI ($\boldsymbol{GHI_{clr}}$), and clear sky index ($\boldsymbol{CSI}$); ii) sky imaging features: mean ($\boldsymbol\mu$), standard deviation ($\boldsymbol\sigma$), and R\'enyi entropy ($\boldsymbol{H}$) of normalized sky image pixel red blue ratio ($\boldsymbol{nRBR}$) values. The feature space is defined as $\boldsymbol{X}=(\boldsymbol{GHI}, \boldsymbol{GHI_{clr}}, \boldsymbol{CSI}, \boldsymbol\mu, \boldsymbol\sigma, \boldsymbol{H})\in \mathbb{R}^{N\times 6}$ ($N$ is the sample length). Data elimination is performed to exclude data in the early morning (before 7am in this study) and late night (after 7pm), since most GHIs are 0. More details of feature extraction can be referred in our previous work~\cite{feng1short}. 

\subsection{Diurnal Patterns and Hourly Similarities}
\begin{figure}[!ht]
	\centering
	\includegraphics[width =3.5in]{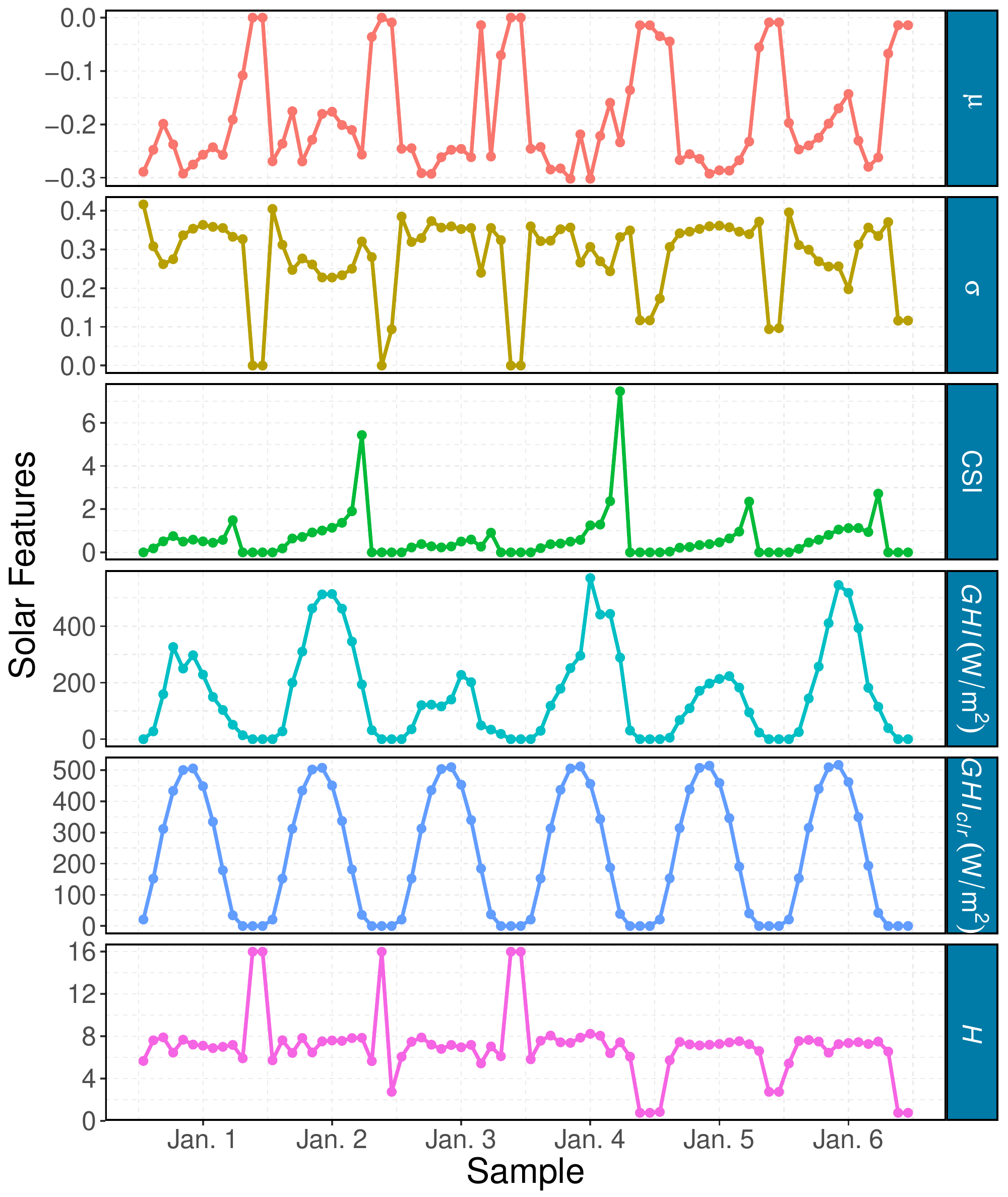}
	\caption{Time series cyclic patterns of solar features.}
	\label{dpattern}
\end{figure}

To ensure the efficiency of HS-based forecasting, diurnal patterns and hourly similarities in solar data need to be explored and identified. Table~\ref{tsc} shows the periodicity, trend, and seasonality of the six solar features, which are calculated by a time series characteristic analysis~\cite{Feng2017Characterizing, wang2006characteristic}. The periodicity in Table~\ref{tsc} indicates the frequency an observation occurs in the time domain. All of the six solar features show \textbf{\textit{diurnal patterns}} (note data elimination was performed, so there are 13 samples in one day). Figure~\ref{dpattern} visualizes the six feature vectors in time series curves, where daily patterns are observed. Also, longer-term patterns also exist in $\boldsymbol{GHI}$, $\boldsymbol{GHI_{clr}}$, and $\boldsymbol{CSI}$ vectors, which are indicated by the trend and seasonality.

\begin{table}[!htb]
	\caption{Time series characteristics in solar features}
	\label{tsc}
	\begin{center}  
		\begin{tabular}{lcccc}  
			\rowcolor{cerulean}&&&
			\\
			\rowcolor{cerulean}\textcolor{white}{\textbf{Feature}}&\textcolor{white}{\textbf{Periodicity}}&\textcolor{white}{\textbf{Trend}}&\textcolor{white}{\textbf{Seasonality}}\\
			\rowcolor{cerulean}&&& \\
			\multicolumn{1}{l|}{$\boldsymbol{GHI}$}&13&0.70&0.84\\
			\multicolumn{1}{l|}{$\boldsymbol{GHI_{clr}}$}&13&0.86&0.96\\
			\multicolumn{1}{l|}{$\boldsymbol{CSI}$}&13&0.22&0.51\\
			\multicolumn{1}{l|}{$\boldsymbol\mu$}&13&0.02&0\\
			\multicolumn{1}{l|}{$\boldsymbol\sigma$}&13&0.06&0\\
			\multicolumn{1}{l|}{$\boldsymbol{H}$}&13&0.16&0.20\\
			\hline
		\end{tabular}
	\end{center}
\end{table}

\begin{figure}[!ht]
	\centering
	\includegraphics[width =3.5in]{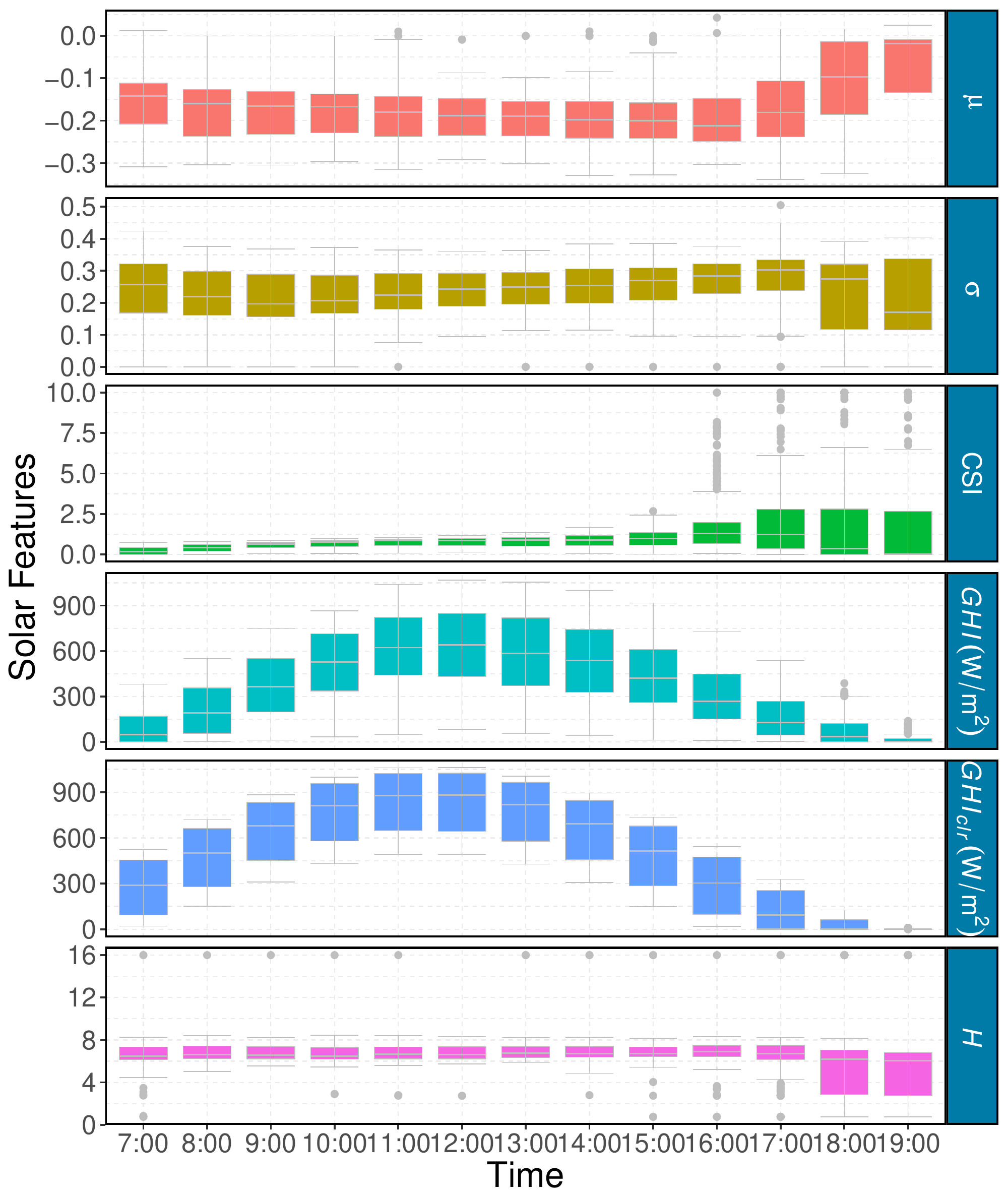}
	\caption{Hourly statistics of solar features. The line in the box is the median. The interquartile range box represents the middle 50\% of the data. The upper and lower bounds are maximum and minimum values of the data, respectively, excluding the outliers. The outliers are data outside two and half times of the interquartile range.}
	\label{boxplot}
\end{figure}

\begin{figure}[!ht]
	\centering
	\includegraphics[width =2.81in]{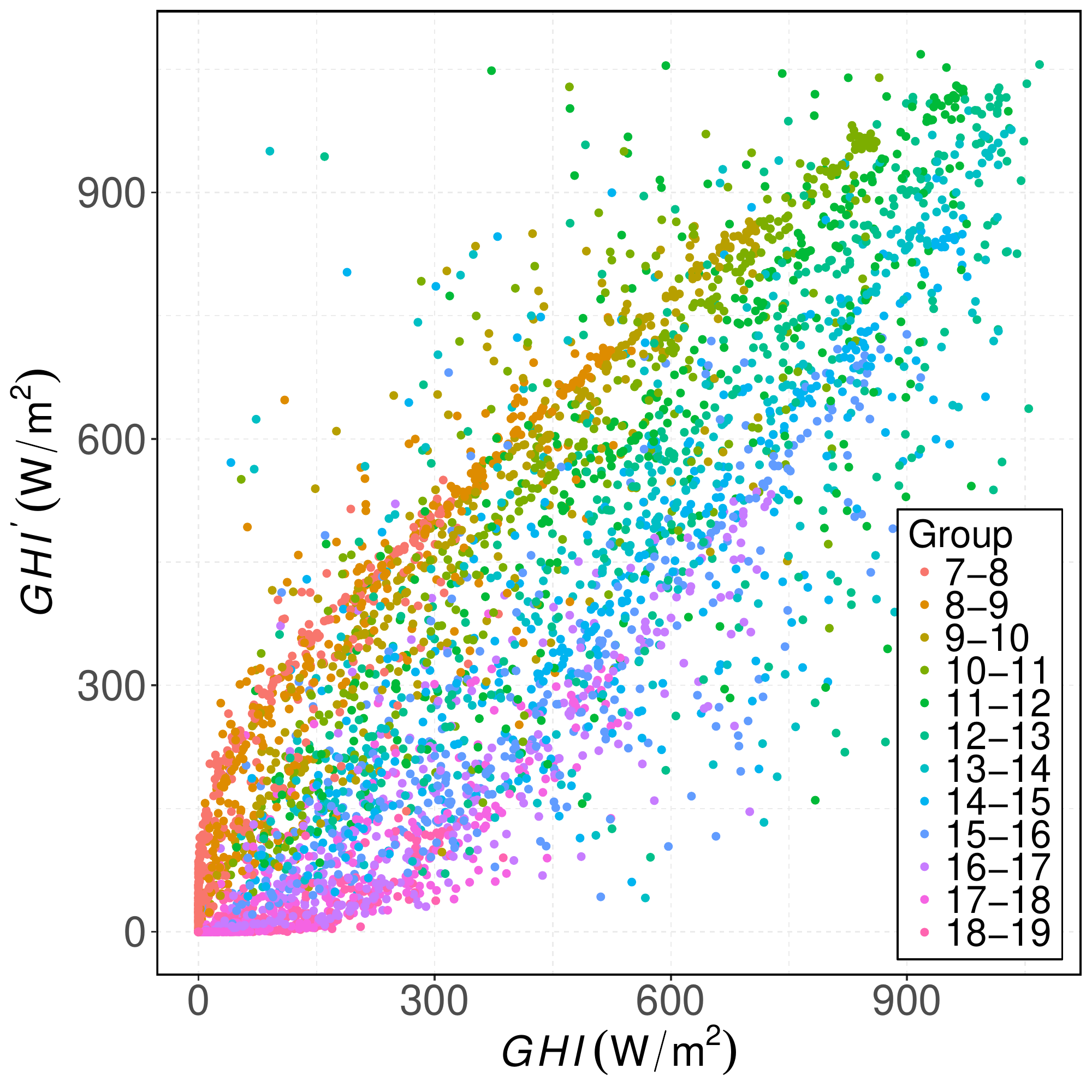}
	\caption{Scatter plot of the current hour GHI ($GHI$) and the 1HA GHI ($GHI'$). Different colors mean different groups. For example, group A-B has the data with current time hour A and next hour B.}
	\label{hourly_scatter}
\end{figure}

Figure~\ref{boxplot} shows the distinctive statistics of solar features in different hours. For example, the median of the same feature in different hours is varying, especially for $GHI$ and $GHI_{clr}$. Moreover, the value range of the same feature in different hours is also different. Thus, solar features in different hours present \textbf{\textit{statistical distinctions}}, and are therefore challenging to be forecasted using a single model. This study focuses on 1HA forecasting, and the scatter plot of the current hour GHI ($GHI_{lag}$) and the 1HA GHI ($GHI$) indicates the most critical relationship between the output and the inputs to forecasting models. Figure~\ref{hourly_scatter} shows a layered pattern with scatter points in the same group dependently converged, which also verifies the \textbf{\textit{hourly similarity}} in solar data. Thus, an HS-based method is expected to improve the the GHI forecasting performance.
\begin{figure*}[tp]
	\centering
	\includegraphics[width =7.2in]{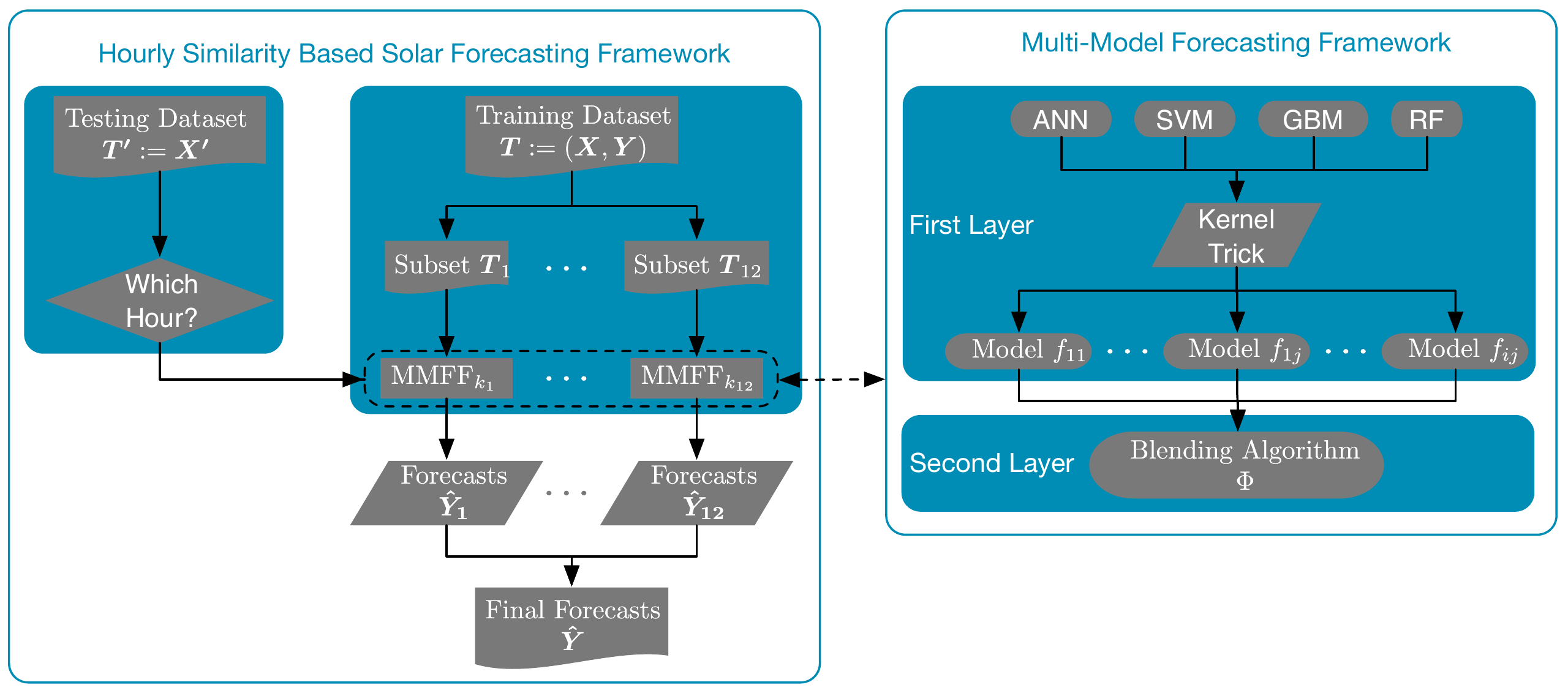}
	\caption{Flowchart of the HS-based GHI forecasting framework with MMFF.}
	\label{flowchart}
\end{figure*}
\section{Hourly-Similarity Based Forecasting Framework Using MMFF}
The flowchart of the developed HS-based forecasting framework using MMFF method is illustrated in Fig.~\ref{flowchart}. The training dataset $\boldsymbol{T} := (\boldsymbol{X},\boldsymbol{Y})$ is hourly partitioned into 12 subsets $\boldsymbol{T}_k~(k = 1,2,...,12)$, each of which is used to train a MMFF model independently. The MMFF model is a two-layer machine learning based method for short-term forecasting. Multiple machine learning algorithms with several kernels generate forecasts, $\tilde{\boldsymbol{Y}}$, individually in the first layer. Then the forecasts are blended by another machine learning algorithm in the second layer, which gives the final forecasts, $\boldsymbol{\hat{Y}}$. Machine learning algorithms used in the MMFF method include artificial neural networks (ANN), SVM, gradient boosting machines (GBM), and random forests (RF). MMFF method has shown to perform better than single algorithm machine learning methods~\cite{feng2017data}. The MMFF method can be expressed as:
\begin{equation} 
\tilde{y}_{t,ij}=f_{ij}(\boldsymbol{x}_{t})
\end{equation}
\begin{equation} 
\hat{y}_t=\Phi(\boldsymbol{\tilde{y}}_{t})
\end{equation}
where $t$ is the time index, $f_{ij}(*)$ is the $i$th first-layer model using kernel $j$, $\tilde{y}_{ij}$ is the forecast provided by the model $f_{ij}$, $\boldsymbol{x}_{t}\in \boldsymbol{X}$ is the input vector to the first-layer models, $\boldsymbol{\tilde{y}}=\{\tilde{y}_{ij}\}$ is the combination of the first-layer forecasts, $\hat{y}_t$ is the final forecast at time $t$, and $\Phi$ is the blending algorithm in the second layer. Note that several blending algorithms can be applied in the second layer, and the best MMFF model, MMFF$_{k_i}$, with a certain blending algorithm in each hour forecasting is selected to constitute the final forecasting framework. This training process is evaluated  through a 10-folder cross-validation in the training dataset. At the forecasting stage, the MMFF model of a certain hour is selected to generate forecasts for that specific hour. The GHIs at 7am are forecasted by a 1-day-ahead (1DA) persistence of cloudiness model, since most GHIs before 7am are close to zero, which do not provide enough information to build an efficient learning model.

\section{Case Study}
\subsection{Data Description and Experimental Setup}
The hourly GHI and sky imaging data released by the National Renewable Energy Laboratory (NREL) was adopted in the case study, which was collected at a location in Colorado (latitude = 39.74$^\circ$ North, longitude = 105.18$^\circ$ West, elevation = 1,828.8 m). The training data was randomly selected from each month, and the remaining data was used for testing. The ratio of the number of training samples to testing samples was 3:1. The experiment was carried out on a laptop with an Intel Core i7 2.6 GHz processor and a 16.0 GB RAM.

Two groups of models were setup for comparison, which were the \textbf{\textit{HS}}-based group and non-similarity based (\textbf{\textit{all-in-one}}) group. In both groups, several machine learning algorithms with multiple kernels were selected as the blending algorithm in the second layer for each forecasting subtask. Details of the kernel information are listed in Table~\ref{kernel}~\cite{feng2018artificial}. In the HS-based group, C$_{opt,h}$ is a combination of multiple MMFF models with the best blending algorithm (i.e., 1DA persistence of cloudiness, RF, SVM$_1$, RF, RF, SVM$_1$, ANN$_1$, ANN$_2$, SVM$_1$, ANN$_3$, RF, GBM$_3$, and RF) for each forecasting subtask. The 1HA persistence of cloudiness model is included in the all-in-one group.

\begin{table}[!htb]
	\caption{Blending algorithm pool for the MMFF second layer}
	\label{kernel}
	\begin{center}  
		\begin{tabular}{clcccc}  
			\rowcolor{cerulean}&&
			\\
			\rowcolor{cerulean} \textcolor{white}{\textbf{Algorithm}}&\textcolor{white}{\textbf{Model Name}}&\textcolor{white}{\textbf{Kernel, learning, or loss function}}\\
			\rowcolor{cerulean}&&\\
			\multirow{7}{*}{\textbf{ANN}}
			&\multicolumn{1}{l|}{}\\
			&\multicolumn{1}{l|}{ANN$_{1}$}&Standard back-propagation\\
			&\multicolumn{1}{l|}{}\\
			&\multicolumn{1}{l|}{ANN$_{2}$}&Momentum-enhanced back-propagation\\
			&\multicolumn{1}{l|}{}\\
			&\multicolumn{1}{l|}{ANN$_{3}$}&Resilient back-propagation\\
			&\multicolumn{1}{l|}{}\\
			\hline	\multirow{5}{*}{\textbf{SVM}}
			&\multicolumn{1}{l|}{}\\
			&\multicolumn{1}{l|}{SVM$_{1}$}&Radial basis function kernel\\
			&\multicolumn{1}{l|}{}\\
			&\multicolumn{1}{l|}{SVM$_{2}$}&Linear kernel\\
			&\multicolumn{1}{l|}{}\\
			\hline	\multirow{7}{*}{\textbf{GBM}}
			&\multicolumn{1}{l|}{}\\
			&\multicolumn{1}{l|}{GBM$_{1}$}&Squared loss\\
			&\multicolumn{1}{l|}{}\\
			&\multicolumn{1}{l|}{GBM$_{2}$}&Laplace loss\\
			&\multicolumn{1}{l|}{}\\
			&\multicolumn{1}{l|}{GBM$_{3}$}&T-distribution loss\\
			&\multicolumn{1}{l|}{}\\
			\hline	\multirow{3}{*}{\textbf{RF}}
			&\multicolumn{1}{l|}{}\\
			&\multicolumn{1}{l|}{RF}&CART aggregation\\
			&\multicolumn{1}{l|}{}\\
			\hline
		\end{tabular}
	\end{center} 
\end{table}

\subsection{Forecasting Accuracy Assessment}
To assess the forecasting accuracy, four evaluation metrics are used, which are normalized mean absolute error ($nMAE$), normalized root mean square error ($nRMSE$), $nMAE$ improvement ($Imp_A$), and $nRMSE$ improvement ($Imp_R$). The mathematical expressions of $nMAE$ and $nRMSE$ can be referred in~\cite{feng1short}. The ($Imp_A$) and ($Imp_R$) metrics are, respectively, defined as:
\begin{equation} 
Imp_A = \frac{nMAE_{M_{l,a}}-nMAE_{M_{l,h}}}{nMAE_{M_{l,a}}}
\end{equation}
\begin{equation} 
Imp_R = \frac{nRMSE_{M_{l,a}}-nRMSE_{M_{l,h}}}{nRMSE_{M_{l,a}}}
\end{equation}
where $M$ is the model name, and $l$ is the kernel index. $h$ and $a$ are group indices indicating the HS-based group and the all-in-one group to which a model belongs, respectively. The focus of this study is to evaluate if the HS-based solar forecasting can improve the accuracy, compared with the all-in-one based forecasting. Therefore, the $Imp_A$ and $Imp_R$ were calculated based on the MMFF models that use the same blending algorithm in the HS-based and all-in-one groups. Note that the $Imp_A$ and $Imp_R$ of the final C$_{opt,h}$ are calculated by comparing with the best model in the all-in-one group. 

\begin{table}[!htb]
	\caption{Overall forecasting performance evaluation}
	\label{forec_accu}
	\begin{center}  
		\begin{tabular}{clcccc}  
			\rowcolor{cerulean}&&&&&
			\\
			\rowcolor{cerulean} \multicolumn{2}{c}{\textcolor{white}{\textbf{Dataset}}}&&&& \\
			\rowcolor{cerulean} \textcolor{white}{\textbf{Group}}&\textcolor{white}{\textbf{Model}}&\multirow{-2}{*}{\textcolor{white}{\textbf{$\boldsymbol{nMAE}$}}}&\multirow{-2}{*}{\textcolor{white}{{$\boldsymbol{nRMSE}$}}}&\multirow{-2}{*}{\textcolor{white}{{$\boldsymbol{Imp_{A}}$}}}&\multirow{-2}{*}{\textcolor{white}{\textbf{$\boldsymbol{Imp_{R}}$}}}\\
			\rowcolor{cerulean}&&&&& \\
			\multirow{10}{*}{\textbf{HS-based}}
			&\multicolumn{1}{l|}{C$_{opt,h}$}&{\textcolor{grn}{\textbf{6.55}}}&{\textcolor{grn}{\textbf{10.07}}}&10.94&7.74\\
			&\multicolumn{1}{l|}{ANN$_{1,h}$}&7.08&10.49&\textbf{11.79}&\textbf{16.13}\\
			&\multicolumn{1}{l|}{ANN$_{2,h}$}&7.20&10.66&\textbf{10.76}&\textbf{14.67}\\
			&\multicolumn{1}{l|}{ANN$_{3,h}$}&7.10&10.51&\textbf{11.54}&\textbf{15.89}\\
			&\multicolumn{1}{l|}{SVM$_{1,h}$}&8.02&12.43&{\textcolor{brickred}{\textbf{-1.32}}}&0.95\\
			&\multicolumn{1}{l|}{SVM$_{2,h}$}&7.54&10.92&4.35&11.72\\
			&\multicolumn{1}{l|}{GBM$_{1,h}$}&7.46&10.87&5.47&12.14\\
			&\multicolumn{1}{l|}{GBM$_{2,h}$}&7.23&10.72&5.40&13.57\\
			&\multicolumn{1}{l|}{GBM$_{3,h}$}&7.45&11.10&4.94&10.24\\
			&\multicolumn{1}{l|}{RF$_h$}&\it\textbf{6.73}&\it\textbf{10.26}&{\textcolor{grn}{\textbf{12.60}}}&{\textcolor{grn}{\textbf{17.04}}}\\
			\hline
			\multirow{10}{*}{\textbf{All-in-one}}
			&\multicolumn{1}{l|}{ANN$_{1,a}$}&8.05&12.51&N/A&N/A\\
			&\multicolumn{1}{l|}{ANN$_{2,a}$}&8.07&12.50&N/A&N/A\\
			&\multicolumn{1}{l|}{ANN$_{3,a}$}&8.02&12.50&N/A&N/A\\
			&\multicolumn{1}{l|}{SVM$_{1,a}$}&8.09&12.55&N/A&N/A\\
			&\multicolumn{1}{l|}{SVM$_{2,a}$}&7.84&12.46&N/A&N/A\\
			&\multicolumn{1}{l|}{GBM$_{1,a}$}&7.89&12.37&N/A&N/A\\
			&\multicolumn{1}{l|}{GBM$_{2,a}$}&\it\textbf{7.64}&\it\textbf{12.36}&N/A&N/A\\
			&\multicolumn{1}{l|}{GBM$_{3,a}$}&7.84&12.37&N/A&N/A\\
			&\multicolumn{1}{l|}{RF$_a$}&7.70&12.37&N/A&N/A\\
			&\multicolumn{1}{l|}{P$_a$}&\textbf{7.35}&\textbf{10.91}&N/A&N/A\\	
			\hline
		\end{tabular}
	\end{center}\textsl{}
	\small Note: The units of all the evaluation metrics are \%. The models with the same model name and numerical subscripts (if any) use the same algorithm. The letter subscript $h$ or $a$ of the models indicates the group it belongs to. 
\end{table}

Table~\ref{forec_accu} lists the forecasting evaluation metrics of all the HS-based models and all-in-one models. The C$_{opt,h}$ model has the smallest forecasting $\it nMAE$ and $\it nRMSE$ among all the models in both groups, as shown in bold green. The P$_a$ model is the 1HA persistence of cloudiness model~\cite{feng1short} that is the best model in the all-in-one group, as shown in bold black in Table~\ref{forec_accu}. The best HS-based method (i.e., $C_{opt,h}$) outperforms the best all-in-one method (i.e., $P_a$) by 10.94\% and 7.74\% based on $nMAE$ and $nRMSE$, respectively. If using a unique blending algorithm in the MMFF models (exclude the $C_{opt,h}$ model) to forecast every hour GHI  (HS-based group), RF$_h$ outperforms other blending algorithm models in the HS-based group. In the all-in-one group, GBM$_{2,a}$ is the best machine learning model, as shown in italic and bold black in Table~\ref{forec_accu}. By comparing the $Imp_A$ and $Imp_R$, the RF$_h$ has the most significant improvement, followed by the ANN models (ANN$_{1,h}$, ANN$_{2,h}$, ANN$_{3,h}$). Compared with all-in-one models, the HS-based models present better performance except for the SVM$_{1,h}$, based on $nMAE$ (as shown in bold red). This SVM model used the radial basis function kernel, which normally needs a large dataset to train. Dividing the training dataset into 12 subsets reduced the sample amount, which possibly caused underfitting during the SVM training. \textbf{\textit{Overall it is found that the HS-based GHI forecasting improved the accuracy significantly}}.

\begin{figure}[!ht]
	\centering
	\includegraphics[width =3.6in]{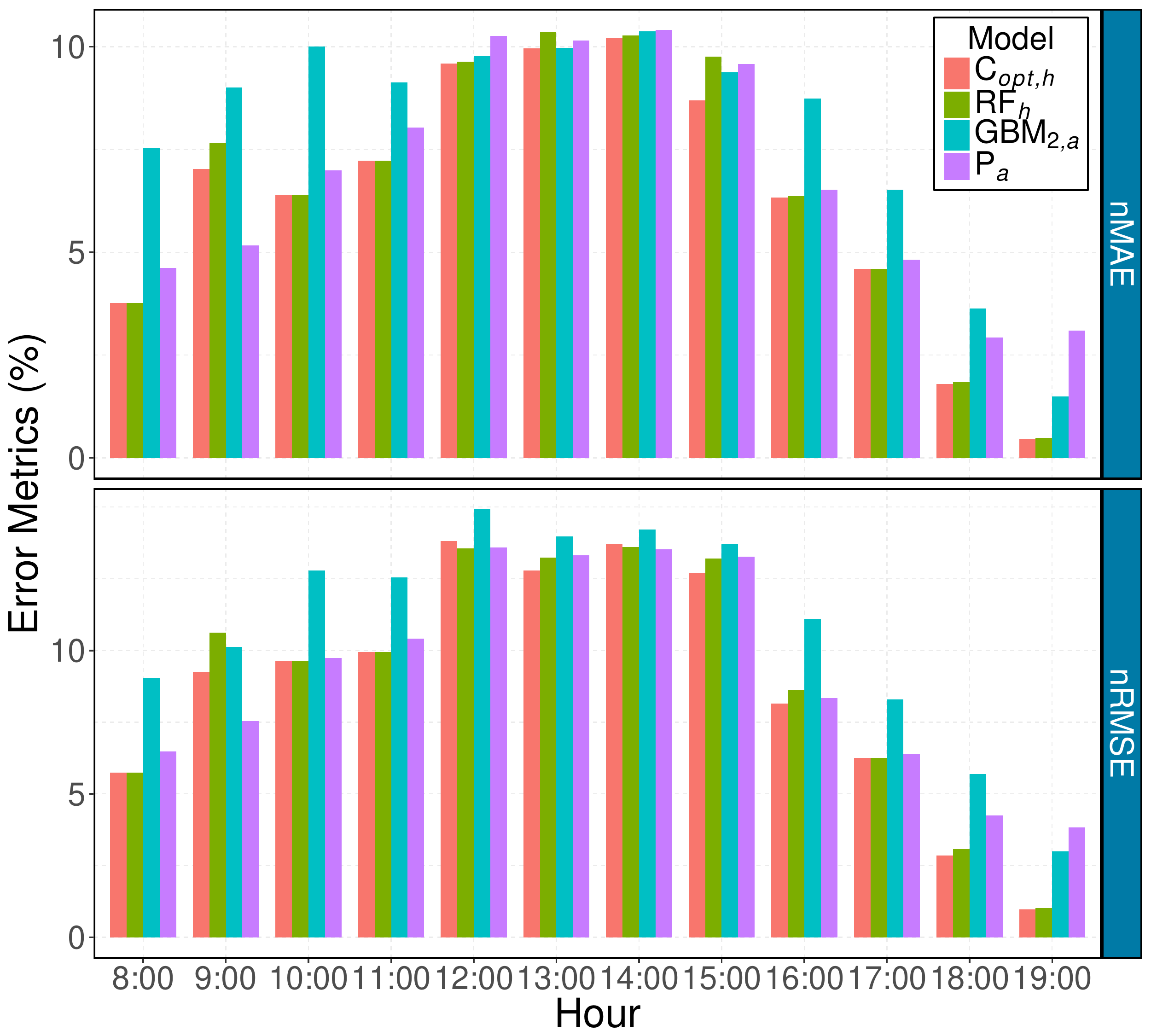}
	\caption{Forecasting errors by hour.}
	\label{error_hour}
\end{figure}

\begin{figure}[!ht]
	\centering
	\includegraphics[width =3.6in]{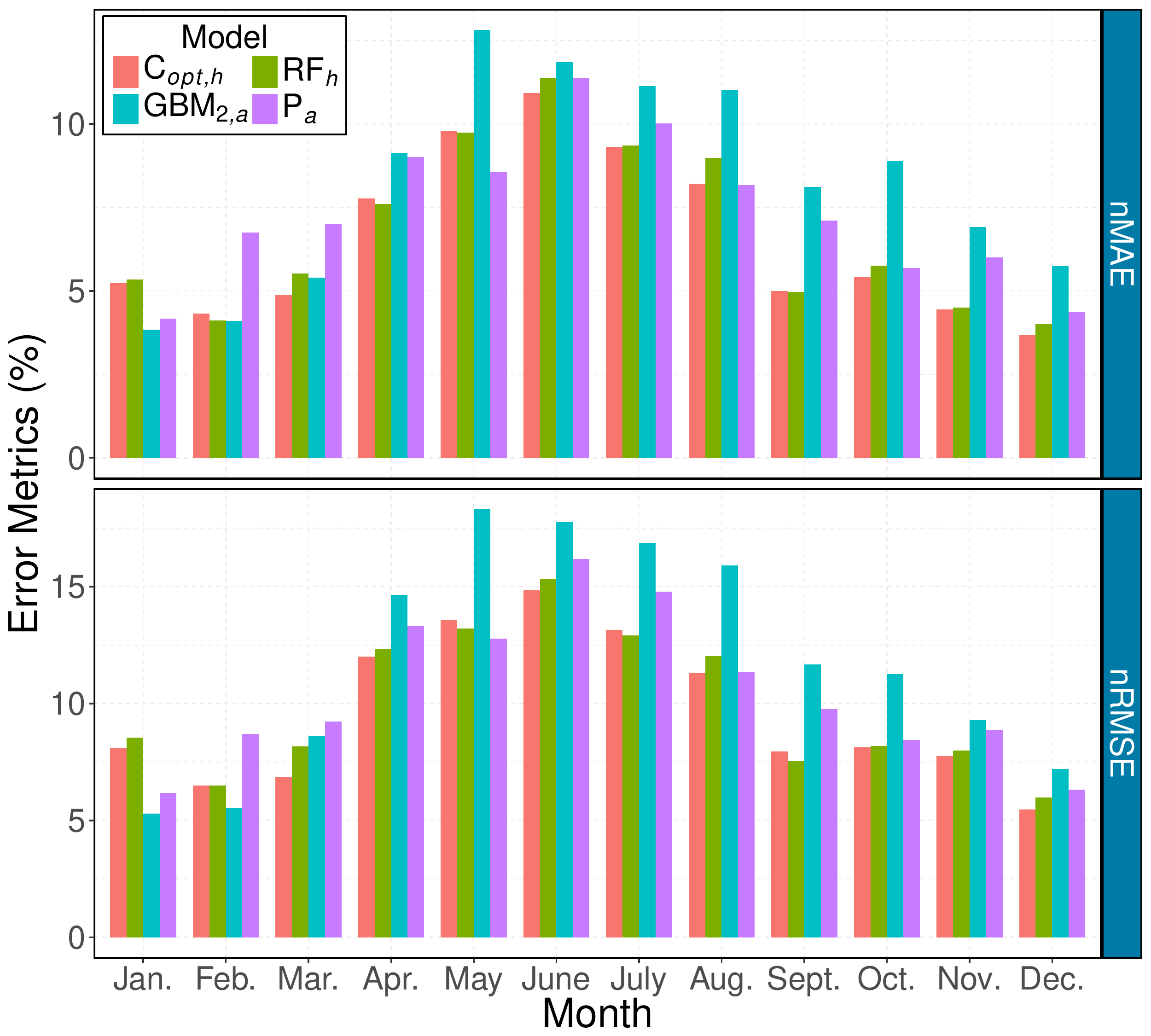}
	\caption{Forecasting errors by month.}
	\label{error_month}
\end{figure}

A further analysis is carried out by comparing the model performance in each hour and each month. Four models are selected, which are the two best models in each group (C$_{opt,h}$ and P$_a$), and two best MMFF models with the unique blending algorithm in the two groups (RF$_h$ and GBM$_{2,a}$). Figure~\ref{error_hour} shows the forecasting errors in every hour. Compared to the methods in the all-in-one group, the HS-based methods have superior forecasting performance, especially when the GHI value is small in the early morning and late afternoon. The stable performance of HS-based forecasts is due to the more stable and similar pattern in the same-hour data. Considering the monthly behaviour, the HS-based methods also perform better than the all-in-one methods, as shown in Fig.~\ref{error_month}. It is seen from Figs.~\ref{error_hour} and~\ref{error_month} that the 1HA persistence of cloudiness model (P$_a$) also shows satisfying performance, considering its simplicity. Thus, P$_a$ method can be used to forecast GHI at some hours, such as 9am, to further improve the forecasting accuracy. The forecasting errors have shown evident diurnal and seasonal patterns, as indicated in Figs.~\ref{error_hour} and~\ref{error_month}, respectively. This information could be used to build seasonal-based models~\cite{vagropoulos2016comparison} or to perform post-processing such as bias correction~\cite{lauret2014neural} in solar forecasting.

\section{Conclusion}
In this paper, an hourly-similarity (HS) based forecasting method was developed to improve the short-term  1HA GHI forecasting. The developed HS-based method forecasted GHI of each hour separately with a two-layer machine learning based MMFF method. Several blending algorithms were used in the second-layer of the MMFF models, the optimal combination of which composes the final optimal HS-based model. A numerical experiment was performed to validate the efficiency by using the 1-year solar data with six features. The developed HS-based 1HA GHI forecasting method has shown superior performance from various perspectives, compared with all-in-one methods.

\bibliographystyle{IEEEtran}
\bibliography{IEEEfull,solarreference3}
\end{document}